# Transferable Above-Ground Biomass (AGB) Estimation Model from Multi-Sensor Data with Sparse Field Calibration

Pann Thinzar Seint, Bryan Atwood, Subas Chhatkuli

*DAI Labs, K.K.*

## ABSTRACT

Continuous, spatially explicit quantification of forest above-ground biomass (AGB) is what makes carbon accounting credible, and mitigation strategies actionable. While field inventories provide high localized accuracy, they are spatially sparse; conversely, spaceborne LiDAR from the Global Ecosystem Dynamics Investigation (GEDI) offers broad biomass samples but lacks wall-to-wall spatial continuity, including systematic underestimation of high-biomass forests. This paper presents an operational framework centered on a single globally trained convolutional neural network (CNN) that is seamlessly adapted to each new landscape through a lightweight empirical field-calibration workflow. The global model combines wall-to-wall optical (Sentinel-2), C-band SAR (Sentinel-1), L-band SAR (ALOS-2 PALSAR-2), and terrain (DEM) data. It is trained once against GEDI Level-4A biomass reference data spanning multiple regions and both wet and dry seasons, allowing it to capture transferable structure–biomass relationships. Training on paired seasonal composites exposes the model to the full annual range of canopy greenness, leaf-on/leaf-off structure, and soil-moisture backscatter conditions, so it learns the persistent woody-structure signal rather than a single-date appearance. Rather than executing computationally intensive retraining for every target landscape, the framework applies a small number of local field plots to fit a scale-and-bias correction that aligns the global prediction with ground truth in each region. The pipeline harmonizes heterogeneous sensors onto a shared 10 m grid, spectral vegetation indices and polarimetric ratios, computes global per-band normalization statistics, and trains the CNN with a hybrid loss that combines a log-domain SmoothL1 term with RMSE to handle the strongly skewed biomass distribution. On held-out validation the global GEDI-based model attains $R^2 \approx 0.78$ and RMSE ≈ 22 Mg/ha, competitive with existing multi-sensor biomass models. A subsequent field calibration combining Random Forest fine-tuning under a 10-fold cross-validation scheme with a final scale-and-bias correction. It eliminates localized regional biases, raising local validation performance to $R^2 \approx 0.82$ and reducing RMSE to ≈ 15 Mg/ha and outperforming both the uncalibrated global model and the ESA CCI Biomass product against field plots.

**Keywords:** *above-ground biomass; global model; field calibration; GEDI L4A; Sentinel-1; Sentinel-2; ALOS-2 PALSAR-2; convolutional neural networks; sensor fusion; wall-to-wall AGB mapping*

# 1. INTRODUCTION

Forest ecosystems play a critical role in the global climate system, sequestering approximately half of all terrestrial carbon and regulating biosphere and atmosphere carbon exchange. Accurate quantification of forest biomass and its temporal dynamics is fundamental to national greenhouse gas inventories, results-based incentives under REDD+, and the monitoring, reporting, and verification (MRV) systems underpinning international carbon markets. REDD+ (Reducing Emissions from Deforestation and forest Degradation) is a framework developed by the United Nations to create incentives for developing countries to protect, manage, and sustainably use their forest resources. By providing financial value for carbon stored in forests, it aims to reduce emissions from deforestation and forest degradation. This framework further fosters conservation, sustainable management of forests, and the enhancement of forest carbon stocks [1]. Aboveground biomass (AGB), the dry mass of living vegetation expressed in Mg/ha serves as the foundational metric within these accounting frameworks. However, achieving accurate, wall-to-wall observations of AGB at regional and global scales remains a critical scientific and operational challenge. A recent study by Zhang et al. [2] further demonstrated the transferability of aboveground biomass estimation models using Sentinel-1/2 and GEDI data in subtropical forests of complex terrain, highlighting the potential for cross-regional model application.

Traditionally, AGB is estimated through destructive harvesting or, more commonly, by applying allometric equations to field-measured tree diameters and heights. While these plot-based methods are precise at the local scale, they are labor-intensive, expensive, and spatially sparse. Consequently, they cannot satisfy the growing demand for spatially continuous, frequently updated biomass maps. Remote sensing provides the only practical approach to achieving wall-to-wall mapping across large, often inaccessible landscapes. Despite its potential, no single remote sensing instrument observes biomass directly. Optical sensors, which measure canopy greenness and structure, frequently suffer from signal saturation in dense forests and are obscured by cloud cover. Synthetic Aperture Radar (SAR) can penetrate clouds and respond to woody structure and moisture; however, radar backscatter is also susceptible to saturation and is confounded by terrain and moisture conditions. Conversely, spaceborne LiDAR provides a physically grounded measurement of vertical canopy structure, but it typically samples the surface as sparse footprints rather than continuous imagery. Since these technologies are highly complementary, sparse LiDAR samples can provide as reference labels to train and calibrate models built on continuous, optical and radar observations.

The Global Ecosystem Dynamics Investigation (GEDI) serves as the foundational source of reference labels for our framework. GEDI is a full-waveform LiDAR instrument installed on the International Space Station, designed to address a critical deficit in Earth observation: the lack of dedicated, global-coverage spaceborne LiDAR for mapping vegetation vertical structure [3]. The GEDI Level-4A (L4A) product offers footprint-level AGBD estimates (Mg/ha) for roughly 25 m footprints. Each estimate is produced by a parametric model that maps waveform relative-height metrics onto biomass, a model calibrated on a large, globally distributed set of field plots paired with airborne LiDAR [4,5]. While GEDI data have already become a cornerstone of vegetation mapping used to calibrate global canopy-height models [6,7] and to train machine-learning models utilizing optical and radar imagery [8,9,10,11], most existing studies rely on retraining models specifically for individual regions. Our work advances this methodology by moving away from region-specific retraining in favor of a more scalable global-model-plus-local-calibration strategy.

Optical imagery is widely used in biomass estimation because canopy reflectance and spectral indices correlate well with canopy density [9,10]. Nevertheless, these indices are inherently limited by signal saturation in dense, high-biomass forests. Synthetic Aperture Radar (SAR) is used to overcome these limitations, as it can penetrate cloud cover and respond to woody structure. Specifically, C-band data are sensitive to smaller canopy elements and moisture, whereas the longer L-band wavelengths penetrate deeper into the canopy to interact with trunks and branches, providing critical data for structurally complex forests [12,13]. Research consistently demonstrates that combining optical imagery with both C- and L-band radar outperforms single-sensor approaches [12,14,15], providing the technical justification for the multi-sensor data fusion stack utilized in this study.

Initial biomass estimation efforts relied on parametric regression and ensemble learners like random forests applied to per-pixel features [8,9,10,15,16]. More recently, Convolutional Neural Networks (CNNs) have advanced this field by learning spatial filters over local image patches, thereby capturing valuable information regarding texture and neighborhood structure [12,14,17]. Despite these advancements, a persistent challenge remains: limited model transferability. Models tuned to one biome often fail to generalize to others, forcing researchers to blend numerous local calibrations to produce continental-scale products [15,16,18,19]. Our approach of training a single, globally robust model that is subsequently calibrated locally with a small set of field plots directly targets this transferability gap. This yields a system that scales like a global model with the scalability and performs like a local one, at the cost of a handful of plots rather than a full training set.

# 2. RESEARCH METHODS

In recent years, deep learning has driven rapid progress in biomass estimation, with new architectures addressing several long-standing problems in the field, including sensor integration, label scarcity, limited transferability, and uncertainty quantification.

Combining information from multiple, complementary sensors has become a central concern. Because no single sensor captures biomass completely, the way in which different inputs are fused strongly affects accuracy. Rather than treating every input equally, attention-based fusion mechanisms learn to weigh each sensor according to how informative it is under a given condition, relying more heavily on radar, for instance, where optical signals become saturated. This selective weighting improves the quality of the features extracted from the multi-sensor stack. Capturing spatial context across large scenes has also received considerable attention. Conventional convolutional networks are restricted to a small local neighborhood and therefore cannot represent relationships between distant regions. Transformer architectures address this limitation by modeling long-range spatial dependencies, which allows the network to relate distant parts of an image and to recognize broad structural patterns across large remote sensing scenes. Extending this idea, multi-modal deep learning frameworks integrate several data sources within a single model, providing the robustness required for reliable carbon quantification at large scales.

A separate challenge is the scarcity of reference labels, which are costly and difficult to obtain. Self-supervised pre-training addresses this problem by first learning representative features from large volumes of unlabeled imagery, so that only a small labeled dataset is subsequently needed to produce accurate wall-to-wall biomass maps. Forest structure and its change over time can now be modeled more explicitly. Graph neural networks represent forest patches as connected nodes and thereby capture the structural connectivity between them, rather than treating pixels in isolation. For change monitoring, high-resolution radar fusion improves the detection of subtle, regionally specific shifts in biomass [25], while temporal consistency constraints reduce noise and stabilize biomass time-series, ensuring that year-to-year estimates remain reliable [26].

Further efforts aim to improve both accuracy and trustworthiness. Multi-scale feature pyramids extract canopy structural information at several spatial resolutions simultaneously, enabling the model to recognize both fine and coarse structural detail [27]. In addition, uncertainty estimation techniques provide confidence intervals alongside each prediction, indicating how far a given estimate can be relied upon and making the outputs suitable for operational use [28]. Deployment across different regions has emerged as a final priority.

Cross-biome transfer learning enables models trained in one ecological zone to be adapted to others, providing the flexibility needed to apply biomass models across diverse and challenging biomes. This objective is closely aligned with the global-model-plus-local-calibration strategy adopted in the present study.

This paper addresses the transferability challenge with a single global model paired with a lightweight field-calibration workflow. In this framework, a Convolutional Neural Network (CNN) is trained once across multiple regions to learn transferable structure biomass relationships from a fused multi-sensor stack. For each new landscape, a small set of local field plots is used to fit a simple scale-and-bias correction, effectively aligning the global prediction to local ground truth. As a result, regions with very limited ground data can still generate high-fidelity biomass maps without the heavy burden of local model development.

# 3. METHODOLOGY

The overall framework is organized into five sequential stages, as shown in **Figure 1**, which outlines the transition from data ingestion to the final corrected AGB map. The first three stages establish a single global model designed to capture transferable relationships between structural features and biomass. The fourth stage utilizes local field plots to adapt this global model to the specific environmental and structural characteristics of a target landscape. Finally, the fifth stage applies the calibrated model to generate spatially continuous biomass maps.

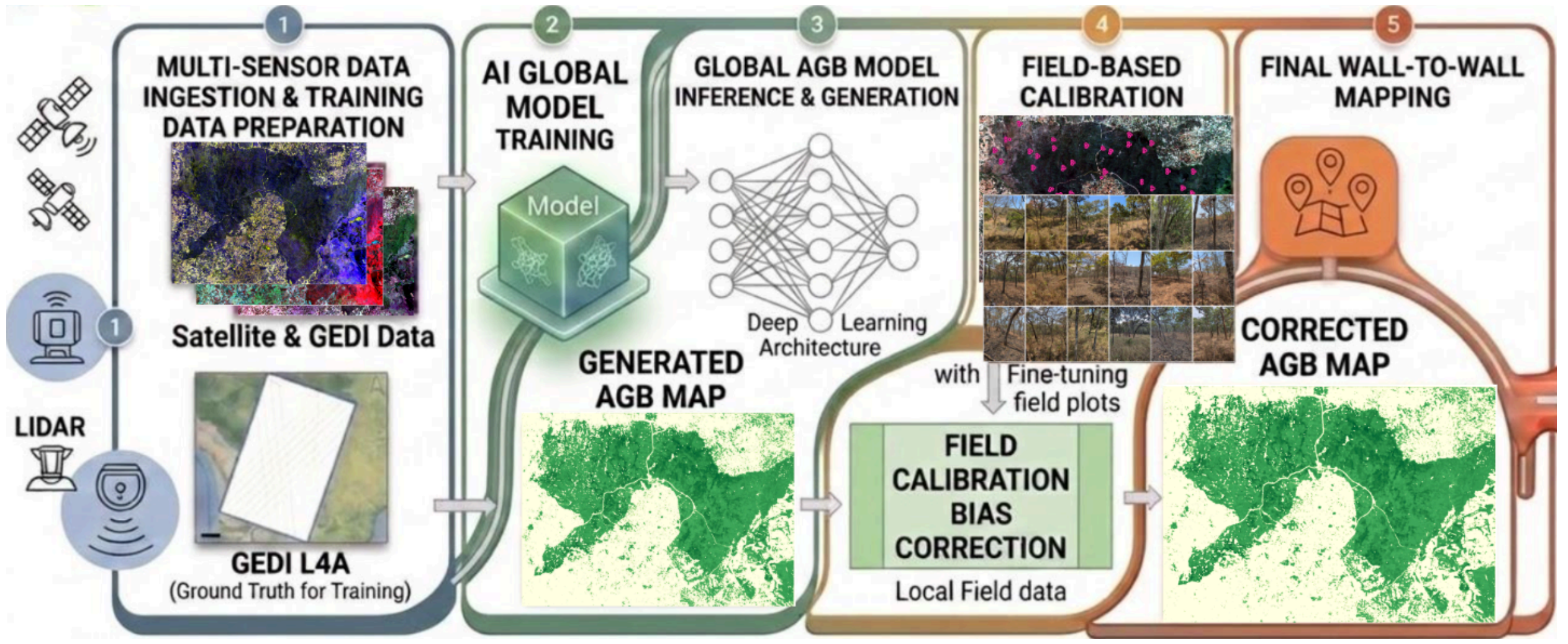

Figure 1. Five-stage system architecture. (Stages 1–3) A global model is trained once and captures transferable relationships; (Stage 4) a lightweight field-calibration step adapts it to each region (Stage 5) wall-to-wall mapping.

## 3.1 Data Integration and Pre-Processing

The framework integrates open source datasets to establish a transferable, operational biomass estimation system. Table 1 details the native resolution and specific role of each data source; importantly, although these inputs vary from 10 m to 30 m in native resolution, the pipeline harmonizes and resamples all data to a consistent 10 m grid for model training and the final wall-to-wall AGB output.

**Table 1**. Data integration used for Global Model

| Data product | Type | Native res. | Description |
|---|---|---|---|
| **GEDI L4A AGBD** | LiDAR | ~25 m | Reference biomass labels (Mg/ha); composited per year |
| **Sentinel-2 L2A (NDVI, NDRE, CCCI, SLAVI, MNDWI)** | Optical | 10–20 m | 12 spectral bands + 5 vegetation indices; canopy greenness and structure |
| **Sentinel-1 RTC** | C-band SAR | 10 m | VV/VH backscatter (dB) + ratio; cloud-tolerant structural signal |
| **ALOS-2 PALSAR-2** | L-band SAR | ~25 m | HH/HV ($\gamma^0$ dB) + ratio; deep-canopy woody structure sensitivity |
| **Copernicus DEM** | Terrain | 30 m | Elevation; defines the reference alignment grid; terrain context |

The primary core datasets and processing operations include:

- **Sentinel-2 (Optical):** Level-2A surface-reflectance scenes are filtered by seasonal windows, low scene-level cloud cover, and 95% area-of-interest (AOI) coverage. Qualifying scenes are ranked by localized cloud fraction and coverage, then median-composited across all 12 bands to suppress residual clouds, shadows, and haze.
- **Vegetation Indices:** Computed directly from the Sentinel-2 reflectance stack to serve as five scale-invariant predictive layers: the Normalized Difference Vegetation Index (NDVI), the Normalized Difference Red Edge index (NDRE), the Canopy Chlorophyll Content Index (CCCI), the Specific Leaf Area Vegetation Index (SLAVI), and the Modified Normalized Difference Water Index (MNDWI).
- **Sentinel-1 (C-band Radar):** Radiometrically terrain-corrected VV and VH scenes are acquired within the same seasonal windows to ensure temporally matched optical–radar pairs. Processing is restricted to a single orbit pass (descending preferred) to prevent backscatter artifacts in steep terrain. The bands are converted to decibels, median-composited, and supplemented with a VV/VH ratio.
- **ALOS-2 PALSAR-2 (L-band Radar):** Dual-polarization HH and HV scenes are converted to gamma-naught decibels and median-composited annually to provide a stable woody-structure baseline against seasonal vegetation dynamics. An HH/HV ratio is appended to the composite; if L-band data are unavailable, the pipeline dynamically falls back to C-band features to maintain operational continuity.
- **Copernicus GLO-30 DEM:** As a Digital Elevation Model (DEM), it represents the continuous topographic surface of the Earth, contributing a critical elevation feature while enforcing spatial uniformity across independently gridded products.

Following spatial alignment with a 10 meter grid, the layers are concatenated into a single 24-band feature tensor comprising 12 Sentinel-2 bands, 5 vegetation indices, 3 Sentinel-1 features, 3 PALSAR-2 features and DEM. To guarantee uniform processing across new regions, every feature band is standardized to zero mean and unit variance using a single set of persisted, globally aggregated statistics rather than per-tile variants.

## 3.2 Dual-Season Framework and Predictor Assembly

A key architectural feature of the global model is its training on two seasonal composites per region and year rather than a single acquisition date. The acquisition windows and the rationale for the dual-season training approach are detailed in Table 2. The year is partitioned into two acquisition windows aligned with the vegetation cycle: a wet-season window spanning growth initiation through peak biomass, and a dry-season window spanning post-harvest drying through leaf drop. Each window yields its own Sentinel-2 and Sentinel-1 composite, and both are paired with the same GEDI biomass reference, so the model sees the same physical biomass under two very different surface appearances.

This dual-season design isolates stable, carbon-storing woody structures from highly variable, transient environmental noise like seasonal canopy greenness, leaf area variations, and moisture-driven SAR backscatter. By training the network on paired wet- and dry-season composites, the model decouples persistent structural signals from temporary phenological shifts, preventing an over-reliance on greenness during wet seasons or a loss of canopy-density data during dry seasons. This approach ensures operational robustness across diverse illumination and moisture conditions, offers deployment flexibility to utilize whichever seasonal window is cloud-free, and doubles the training sample size via genuine physical variation instead of synthetic augmentation, ultimately mitigating seasonal regional biases prior to field calibration.

**Table 2**. Seasonal acquisition windows. Each region–year contributes both composites, paired with the same GEDI biomass reference, so the model observes identical biomass under contrasting phenological and moisture conditions.

| Season | Acquisition window | Vegetation state and what the model learns |
|---|---|---|
| **Wet** | December – May | Growth initiation through peak biomass; full canopy, high greenness, wet soils. Supplies canopy-density and leaf-area information. |
| **Dry** | June – November | Post-harvest drying through leaf drop; reduced greenness, drier soils. Exposes woody structure with less leaf and moisture confounding. |

## 3.3 Global-Model Training and Inference

At the core of our framework is a Convolutional Neural Network (CNN) designed to translate complex multi-sensor imagery into precise biomass estimates. The architecture uses a two-stage convolutional process to extract spatial patterns: initial layers identify simple textures, while deeper layers capture sophisticated structural features. Adaptive pooling then distills these features into a compact descriptor for a fully connected head, which produces the final AGB prediction. A final softplus activation keeps all outputs non-negative and smooth, consistent with the fact that biomass cannot be negative.

Biomass is not a per-pixel property: canopy texture, gaps, and structural heterogeneity are expressed across neighbourhoods, not single cells. Rather than treating pixels in isolation, the model exploits local spatial context. Each training sample is a 5 × 5 × 24 patch: a twenty-five-pixel neighbourhood observed across all twenty-four bands  centred on a pixel carrying a valid GEDI biomass label. In total, the model was trained on 281,872 patches and evaluated on an independent set of 74,388 validation patches.

The training (70%) and validation (30%) splits are executed within each individual region/season/year tile before being combined; this structured validation strategy ensures that performance metrics reflect true operational domain behaviour across all tiles, seasons, and years rather than a localized global draw. To handle the highly right-skewed distribution typical of forest biomass data, the network is optimized using a hybrid loss function. This objective combines a log-domain Smooth L1 term which stabilizes gradient updates in high-biomass, old-growth forests with a standard Root Mean Squared Error (RMSE) term that maintains absolute precision across low-to-moderate biomass ranges. This balance of multi-scale spatial features and robust statistical constraints allows the network to learn generalized structural relationships. The resulting global GEDI-based model demonstrates strong and well-calibrated performance on held-out data, achieving **$R^2$ ≈ 0.78** and **RMSE ≈ 22 Mg/ha** across the full validation set. The corresponding predicted-versus-actual AGBD alignment is illustrated in **Figure 2**. **Figure 3** illustrates the residuals, calculated as the difference between predicted and actual AGBD values. To reveal the residual trends across the biomass range, residuals are grouped into 10 Mg/ha bins, and the mean residual is computed for each bin. Positive values indicate model overestimation (predictions higher than actual), while negative values indicate underestimation (predictions lower than actual).

At the Inference phase, the trained model is deployed in a wall-to-wall inference pass using the identical 5 × 5 × 24 neighbourhood formulation. Whereas GEDI provides biomass only at discrete footprints, inference generalizes these labels to every pixel. For every pixel in a scene, the surrounding 5 × 5 window is extracted and the model estimates the biomass of its central cell, yielding spatially continuous biomass maps with complete coverage across each region.

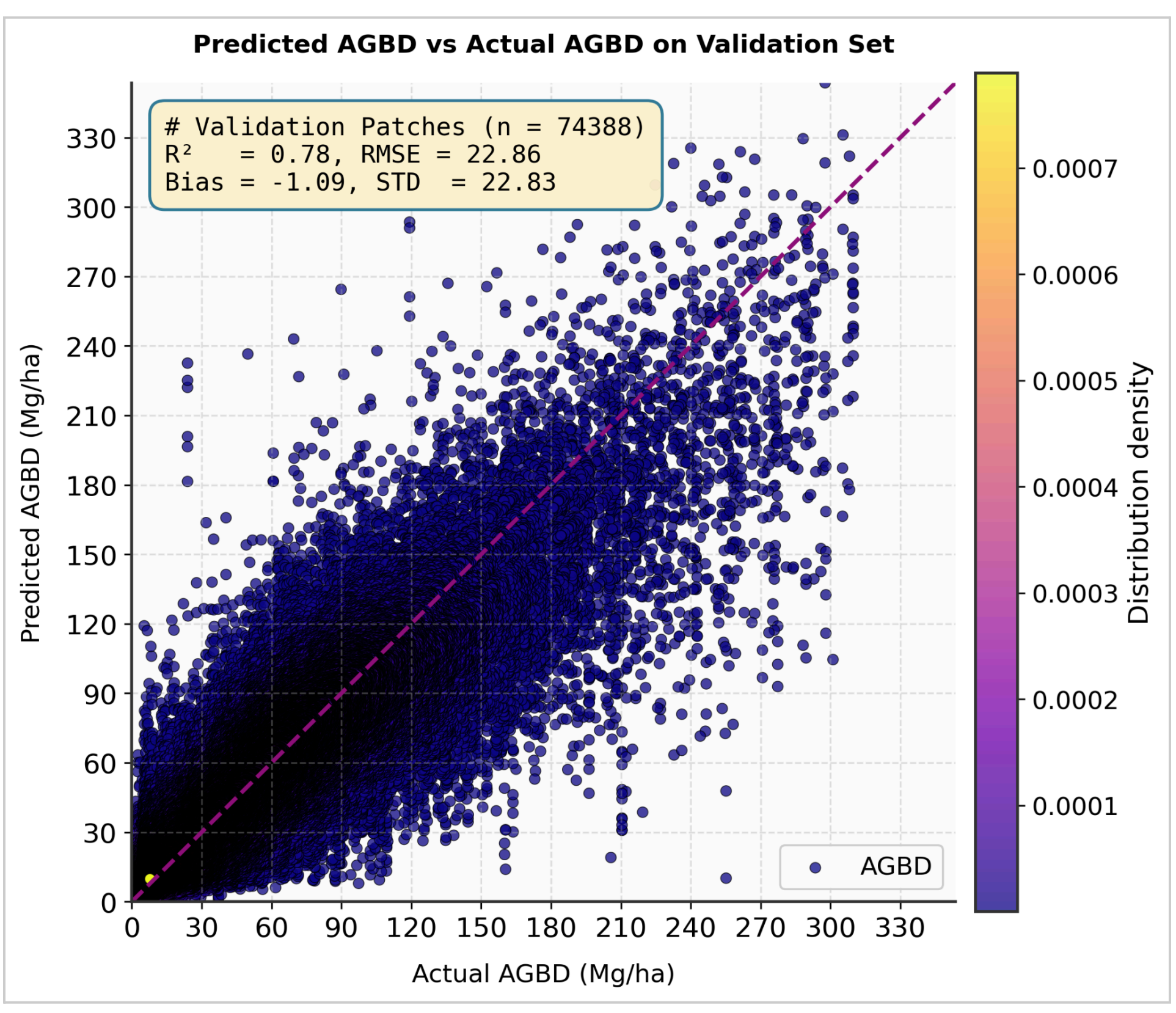

Figure 2. Predicted AGBD vs Actual AGBD on Validation Set. The GEDI-based AI global model demonstrates high correlation ($R^2$ ≈ 0.78) across the diverse validation dataset.

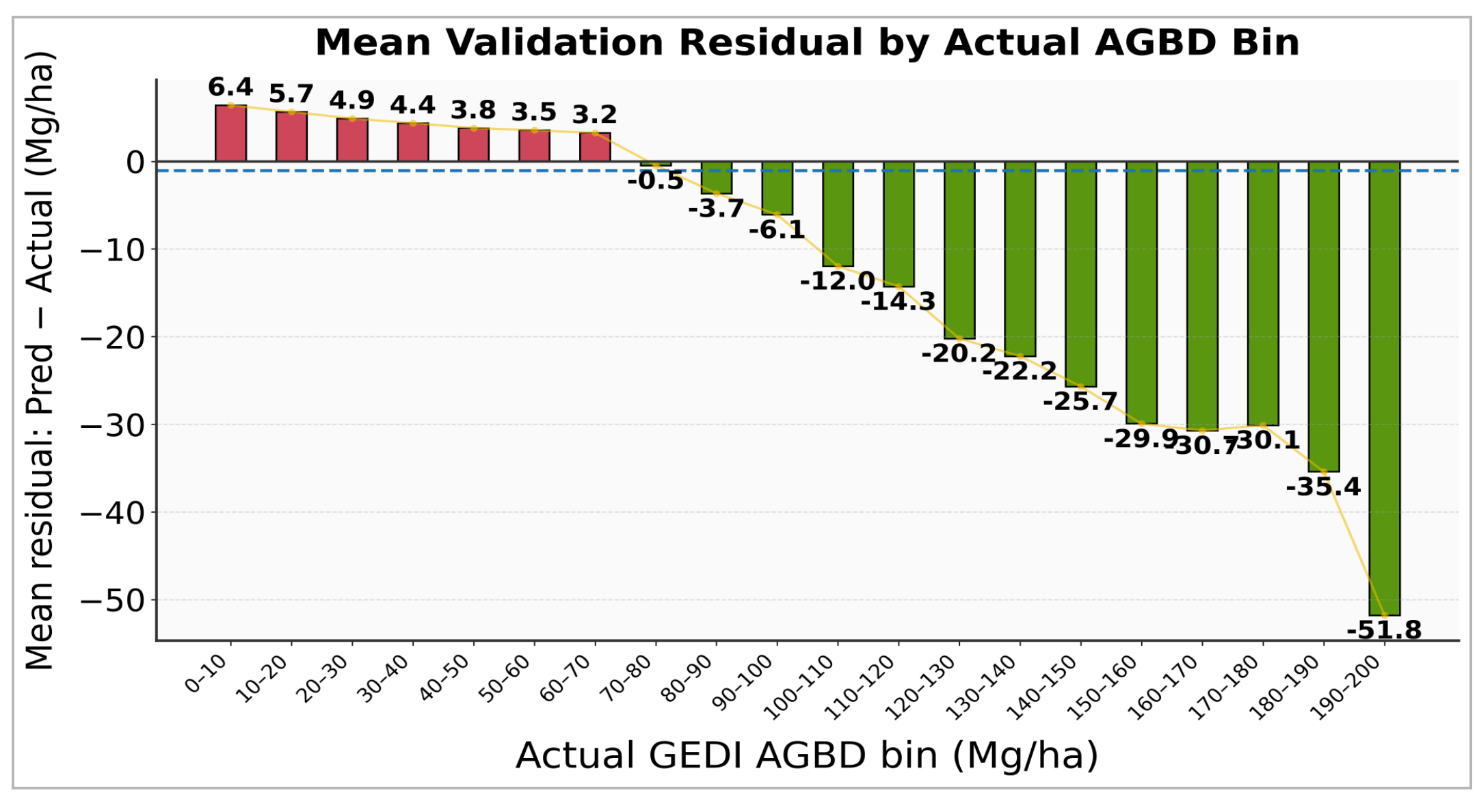

Figure 3. Mean residual between predicted and reference GEDI AGBD across 10 Mg/ha bins.

### 3.4 Field Measurement of Above-Ground Biomass (AGB) Using Nested Circular Plots: Perekezi Forest, Malawi, 2025

We use nested circular plot measurements for quantifying above-ground biomass (AGB) at the Perekezi Forest Reserve in 2025, collected via a KoboToolbox form. Each cluster comprises four plots in a "T" configuration: a central plot (C) with north (N), east (E), and south (S), each located 150 m from the center as shown in **Figure 4**. Trees are recorded in nested concentric subplots with increasing DBH thresholds: 1–<10 cm within 5 m, 10–<20 cm within 10 m, 20–<30 cm within the 15 m (primary) subplot, and ≥30 cm within the 20 m (upscale) subplot, so that abundant small stems are sampled intensively while rare large trees are captured across a wider area. For each tree, species, DBH, total height, crown dimensions, status, and origin are recorded; plot metadata include GNSS location, slope, aspect, canopy closure, disturbance, and photographs.

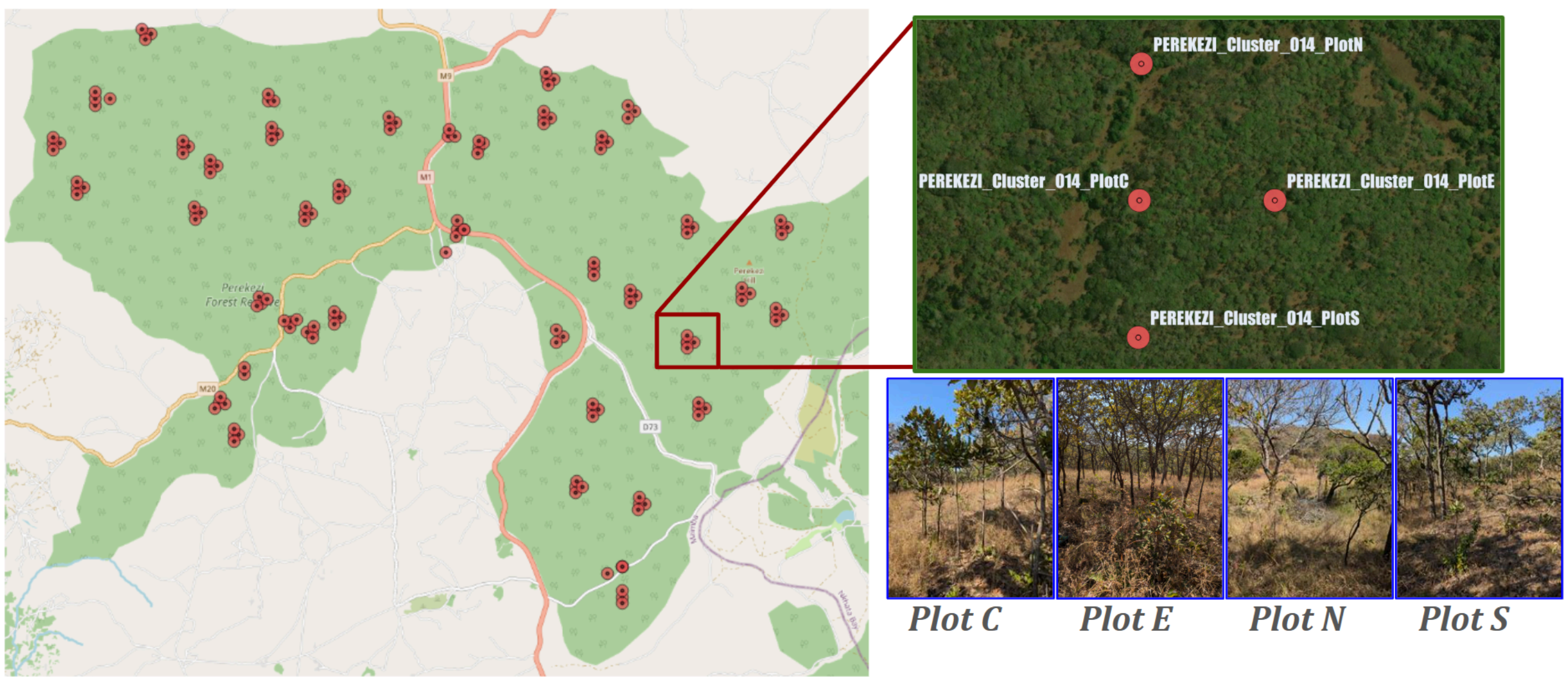

Figure 4. Field data collection design for above-ground biomass measurement using nested circular plots at Perekezi Forest Reserve, Malawi (October–November 2025).

Then, above-ground biomass is calculated using the Kachamba et al. (2016) allometric equation [30] which was developed specifically for the miombo woodlands of Malawi. This equation is adopted in place of generic pan-tropical allometric models because it was calibrated on locally destructively-sampled trees spanning the DBH and height ranges typical of Malawian miombo, and therefore reflects the species composition, wood density, and tree architecture of the study area. Its use improves the accuracy of AGB values for Perekezi and reduces the bias that can arise when applying equations derived from other regions or forest types. The resulting plot-level AGB values provide ground-truth reference data for calibrating a global biomass model to local conditions in Perekezi Forest Reserve.

## 3.5 Field-Calibration Workflow

The overall processing architecture is structured into two independent stages, separating computationally intensive global feature extraction from localized model adaptation (**Figure 5**). Part A consists of a one-time global training phase where multi-region satellite imagery and GEDI labels are assembled into normalized feature patches to train the base Convolutional Neural Network (CNN). Part B includes the local adaptation framework, which ingests the initial global predictions alongside auxiliary remote sensing layers to execute a localized refinement pipeline using regional field plots.

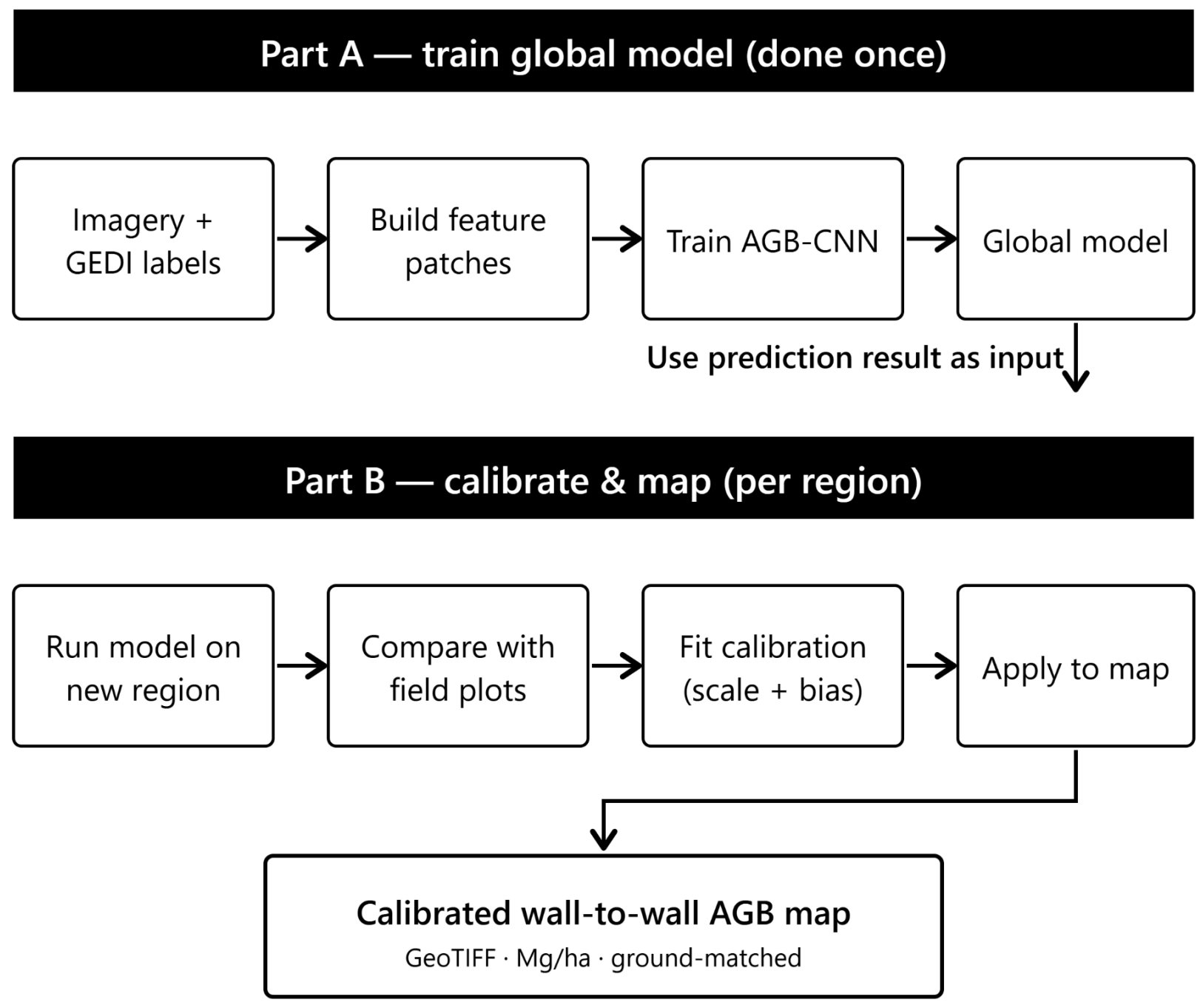

Figure 5. Global-model training (Part A, once) and per-region field calibration (Part B).

We used a minimum of 50 field plots per site, collected by local communities in Malawi and Forest Research Institute of Malawi. These field data are essential for correcting regional bias in the global model and producing biomass maps suitable for carbon accounting. We calibrated the model into  localized adaptation using field measurement plots as targets, with auxiliary satellite imagery layers and GEDI-based predicted AGB map as input features and then bias-corrected. The localized adaptation is executed sequentially through two consecutive phases: a **Random Forest Ensemble Fine-Tuning** stage to model non-linear regional variations, followed by a **Polynomial Bias Correction** step to eliminate systematic scale and offset errors.

### 3.5.1 Random Forest Ensemble Fine-Tuning

To adapt the generalized global model representations to a specific landscape, a localized machine learning refinement pipeline is implemented through the following sequential steps:

1. **Feature Stack Assembly:** The pixel-level predictive map from the global model (GEDI-based AGB map) is stacked together with all co-registered satellite rasters, including Sentinel-1, Sentinel-2, ALOS-2 PALSAR-2, and DEM, to form a comprehensive local input feature space.
2. **Spatial Neighborhood Feature Extraction:** To reconstruct localized structural discrepancies and mitigate pixel-level registration noise, 7*7 moving windows are applied across the entire stack. This operation extracts spatial neighborhood mean values across all input features.
3. **Field-Plot Sample Extraction:** The window-averaged features are spatially sampled at the exact geographic coordinates of the local field plots, pairing the multi-sensor features with the ground-truth biomass measurements to construct the field-plot dataset.
4. **Ensemble Training via k-Fold Cross-Validation:** A Random Forest regressor is trained using a 10-fold cross-validation scheme as shown in **Figure 6**. Random Forest was chosen for its robustness with a small dataset which builds an ensemble of regression trees, each grown on a bootstrap sample with a random feature subset at every split, and averages their predictions to reduce variance and suppress overfitting. The ensemble uses 400 trees, the squared-error criterion, a maximum depth of five, and square-root feature selection per split to constrain complexity on the small field-plot dataset. Within each individual fold, the extracted field-plot dataset is split into 70% for training and 30% for validation. The model evaluates out-of-fold performance for each iteration and subsequently projects a wall-to-wall predictive biomass map for that fold.
5. **Composite Mapping:** The resulting ten wall-to-wall AGB maps are averaged pixel-by-pixel to produce a single, cross-validated mean AGB map. Averaging across folds further suppresses fold-specific variance and yields a stable prediction, which serves as the base input for the final bias-correction stage.

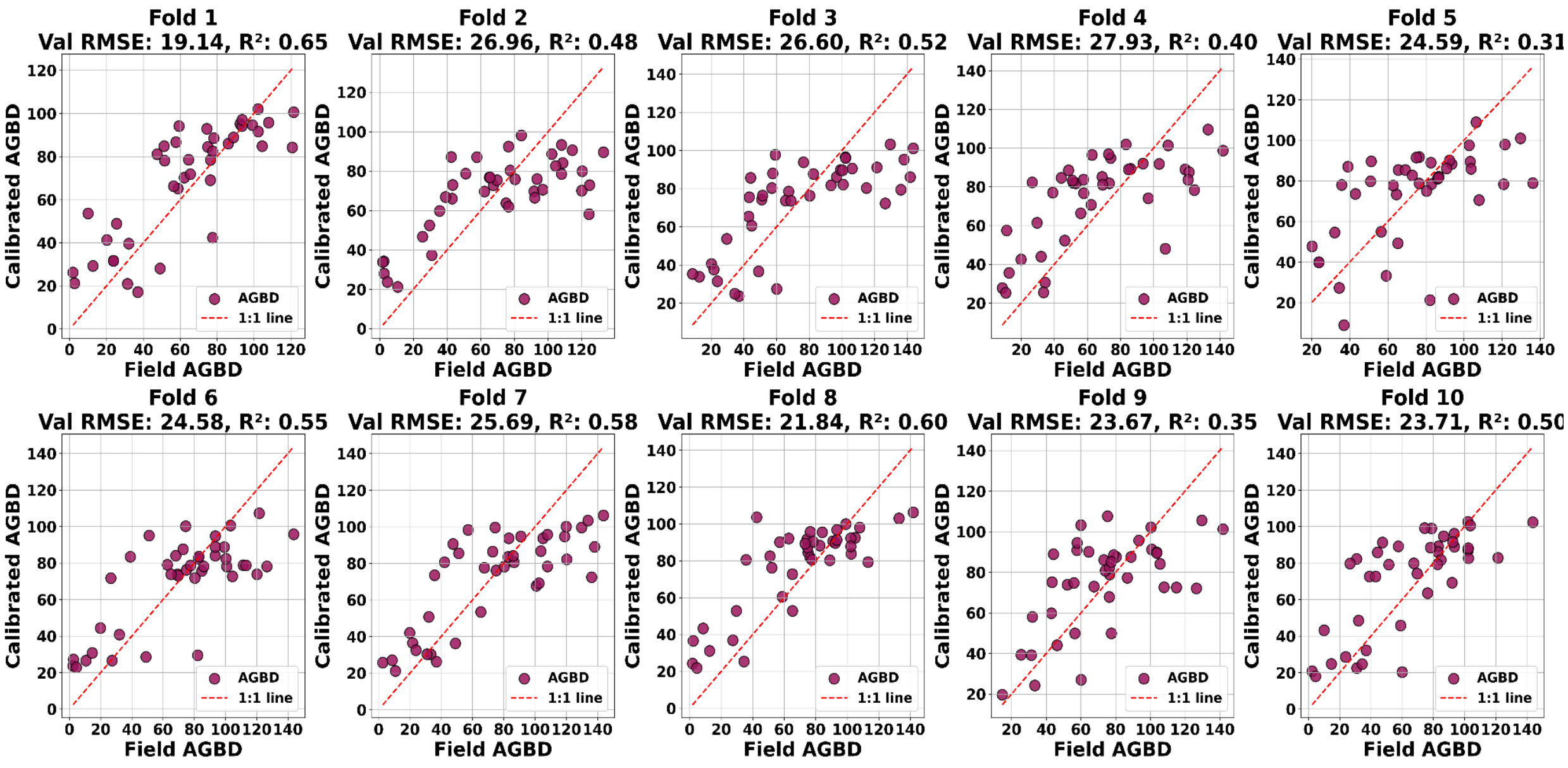

Figure 6. K-Fold cross-validation results for the Random Forest ensemble fine-tuning.

### 3.5.2 Polynomial Bias Correction

To eliminate systematic regional offsets caused by local species composition, structural allometries, an empirical bias-correction function is applied to the fine-tuned ensemble map. The initial calibrated and cross-validated mean AGB map is first sampled at the coordinates of the 149 regional field plots surveyed in 2025 at Perekezi Forest Reserve, Malawi.  to yield paired observations of predicted and field-measured biomass. This dataset is randomly partitioned, reserving 70% of the plots for calibration and 30% for independent validation of the correction. Using the calibration subset, a third-order polynomial regression is fitted to mathematically map the fine-tuned model outputs to the ground truth. The bias-correction function is defined as:

$$AGB_{cal} = aAGB^{3}_{pred} + bAGB^{2}_{pred} + cAGB_{pred} + d$$

where, $AGB_{pred}$ represents the cross-validated mean AGB derived from the fine-tuned Random Forest ensemble, $AGB_{cal}$ is the final calibrated, bias-corrected aboveground biomass estimate Mg/ha, and *a*, *b*, *c*, and *d* denote the polynomial calibration coefficients empirically estimated from the training data. This higher-order function scales the predictive values across the entire biomass gradient, directly mitigating systematic over- or under-estimation before final map generation.

# 4. RESULTS AND ACCURACY

A global model minimizes average error across many regions, but any individual landscape can carry a systematic offset used by local species composition, soil and moisture background, or GEDI sampling density. Field calibration corrects exactly this systematic component. By regressing a small set of field-plot AGB values against the co-located global-model predictions, this removes region-specific bias and improves absolute accuracy where it matters for carbon accounting, while leaving the transferable structure learned by the global model intact.

The model performance is quantified across both the training and held-out validation sets at each epoch using four complementary statistical metrics: the coefficient of determination, root-mean-square error, the mean bias, and standard deviation. On the held-out validation set drawn from the per-tile stratified split, the global GEDI-based model demonstrates strong predictive performance across a heterogeneous, multi-region dataset, achieving an R2 0.78, RMSE 22.86 Mg/ha and a negligible systematic offset Bias = -1.09 Mg/ha with a residual error standard deviation of 22.83 Mg/ha.

However, testing the global model directly against the localized 2025 field plot inventory from Perekezi forest reserve in Malawi highlights the necessity of regional adaptation. When validated against these local field measurements, the uncalibrated global GEDI-based model yields a lower correlation **R2 = 0.11** and an **RMSE = 33.5 Mg/ha**, capturing broad structural trends but failing to resolve local landscape variations. Implementing the localized Random Forest fine-tuning and polynomial calibration workflow dramatically improves local accuracy. The fully calibrated framework resolves these regional discrepancies, elevating the local validation performance to an **R2 of 0.82** and halving the prediction error to an **RMSE of 15 Mg/ha**. A detailed comparison of these performance metrics across Perekezi, Ntchisi, and Dzalanyama contrasting the uncalibrated global base model against the fully calibrated framework is presented in **Table 3**. This improvement demonstrates that while the global architecture successfully learns structural representations, the localized calibration phase is essential for delivering highly accurate, policy-grade biomass maps for specific target regions.

**Table 3**. The Performance of GEDI-based Model and Local Field Calibration

| Regions | Global-GEDI Model (10 meter) | | After Field Calibration (10 meter) | |
|---|---|---|---|---|
| | R2 | RMSE (Mg/ha) | R2 | RMSE (Mg/ha) |
| Perekezi (2025) | 0.11 | 33.58 | 0.82 | 15.00 |
| Perekezi (2020) | -0.85 | 41.88 | 0.83 | 15.51 |
| Ntchisi (2022) | -2.1 | 50.58 | 0.80 | 12.82 |
| Dzalanyama (2020) | 0.16 | 27.28 | 0.82 | 12.55 |

The calibrated AGB is for final delivery, wall-to-wall biomass map. Representative field plots and scatter results of this calibration process across different study regions showing the transition from global model predictions to ground-matched output are displayed in **Figure 7**. We further benchmark our results against the ESA Climate Change Initiative (CCI) Biomass product, a global aboveground biomass dataset produced by the European Space Agency from radar missions such as Envisat ASAR, ALOS PALSAR, and Sentinel-1. Released in successive versions since 2018, it has become a widely used reference for large-scale carbon monitoring, providing global maps at roughly 100 m resolution, though its coarse scale and global calibration limit accuracy over small, heterogeneous landscapes. As shown in **Figure 7 (b)**, validating the three approaches of (i) the ESA CCI Biomass product, (ii) the uncalibrated GEDI-based model, and (iii) our ground-calibrated framework against the local field plots reveals that our calibrated framework agrees most closely with the measured biomass.

**Figure 8** compares field-measured AGBD against predictions from the GEDI-based model and three field-calibrated variants (without ALOS, with 10 m ALOS, and with 6 m ALOS) for the Perekezi site in 2025. Incorporating ALOS PALSAR-2 data calibration visibly tightens the agreement between predicted and field-measured plot values relative to the uncalibrated GEDI-based model and calibrated model without ALOS, reflected in an improved $R^2$ and lower RMSE. At 10 m resolution ALOS data combination, **$R^2$** rises from **0.79 to 0.82**, while the 6 m variant yields a further improvement. Calibrated wall-to-wall AGB maps are shown in **Figure 9.**

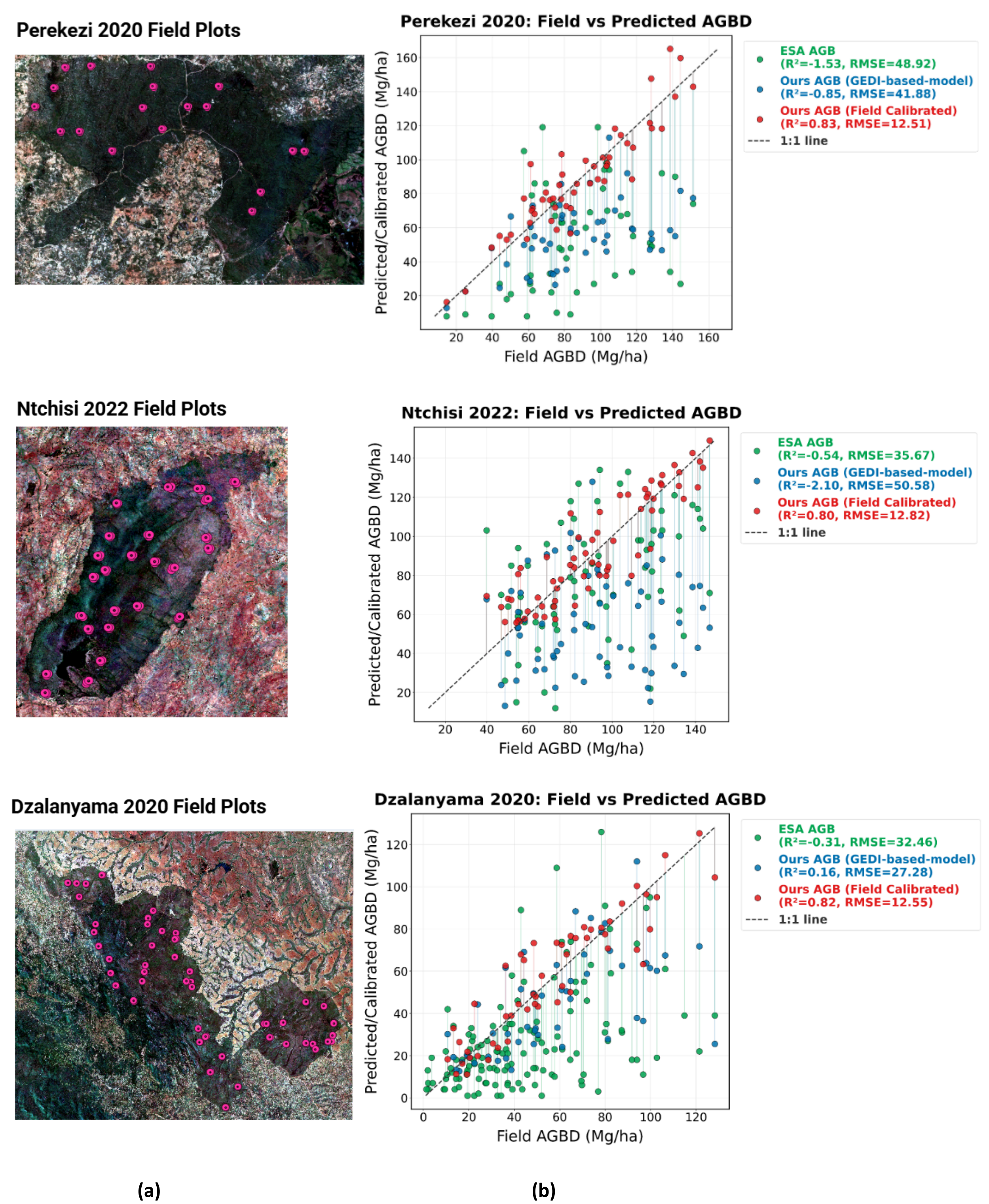

Figure 7. Regional field calibration results for Perekezi (2020), Ntchisi (2022), and Dzalanyama (2020): (a) Location of field plots, (b) Comparison between (i) ESA Biomass, (ii) GEDI-based model predictions and (iii) Field calibration against field data.

Figure 8. Effect of ALOS PALSAR-2 incorporation and resolution on field-calibration accuracy: predicted vs. field AGBD (Mg/ha) for the GEDI-based model and three calibration configurations, Perekezi 2025.

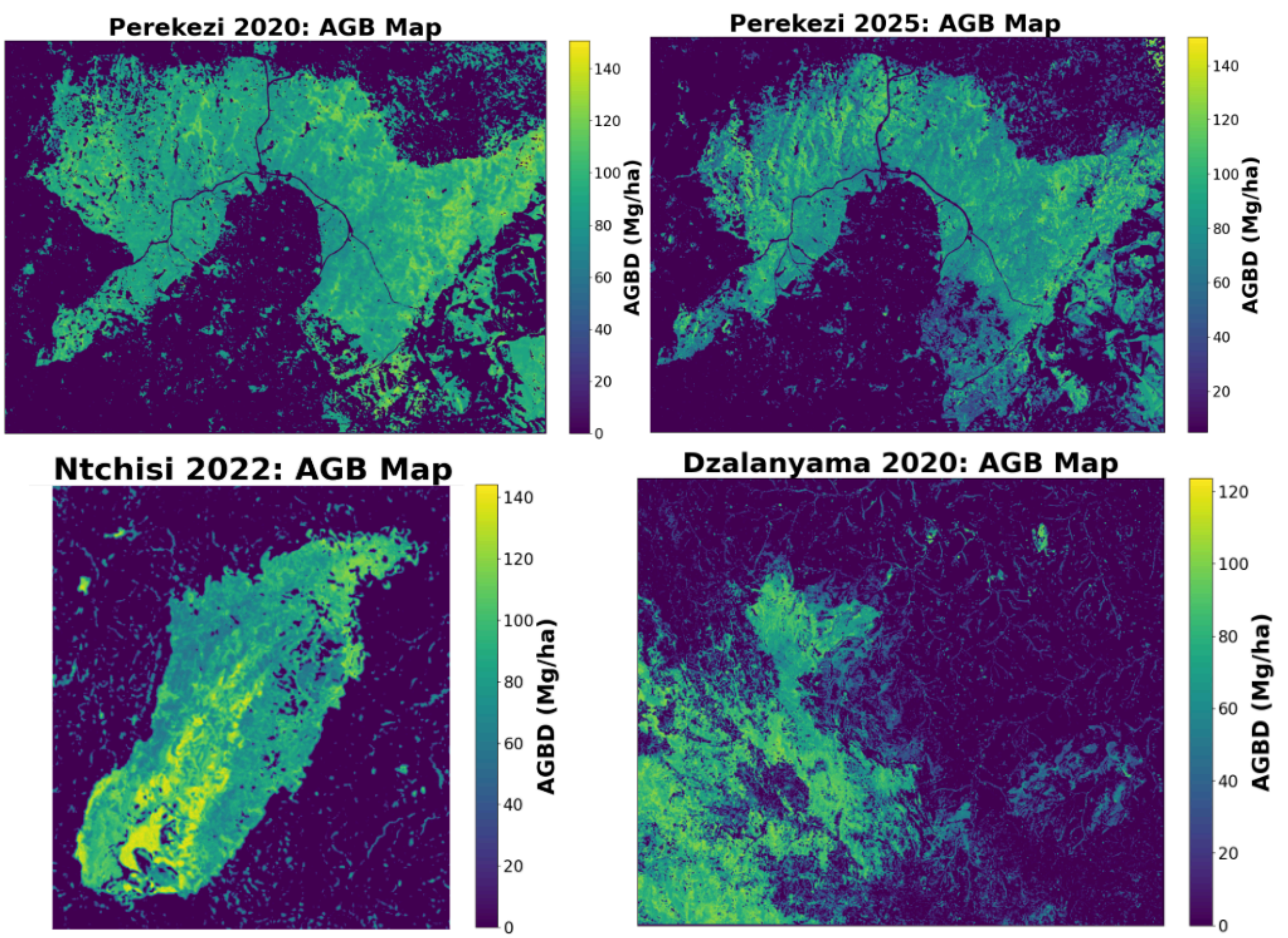

Figure 9. Field Calibrated AGB Maps for Perekezi 2020, Perekezi 2025, Ntchisi 2022 and Dzalanyama 2020.

# 5. APPLICATIONS AND OPERATIONAL VALUE

By transforming publicly available satellite observations into ground-truth-anchored, accurate wall-to-wall biomass maps, this framework provides a highly scalable solution across environmental, policy, and commercial domains as shown in **Figure 10**. The separation of global structural learning from localized field calibration ensures that these spatial products are highly accurate and easily portable across multiple operational applications.

- **Carbon Accounting and Climate Policy:** Supplies the high-resolution activity data required by MRV protocols to establish carbon baselines and emission factors. The Random Forest refinement with calibration pipeline eliminates regional systematic offsets, ensuring compliance-grade confidence for national GHG inventories, REDD+, and voluntary markets [11,19].
- **Forest Management, Conservation, and Restoration:** Optimizes sustainable-yield planning, identifies high-conservation-value assets, and prioritizes strategic ecological restoration sites. Continuous monitoring enables rapid detection of forest degradation and illegal logging while verifying long-term carbon sequestration goals [18,31].
- **Ecosystem, Biodiversity, and Disaster Applications:** Leverages structural biomass density as a direct proxy for habitat quality, biodiversity assessments, and ecosystem service valuations. It quantifies available forest fuels for wildfire modeling.
- **Agricultural and Infrastructure Uses:** Optimizes inventory management, growth rate assessments, and yield predictions for commercial timber plantations and agroforestry systems. Furthermore, it strengthens vegetation hazard assessment models for crop insurance underwriting and guides right-of-way maintenance along high-voltage utility and transport corridors.

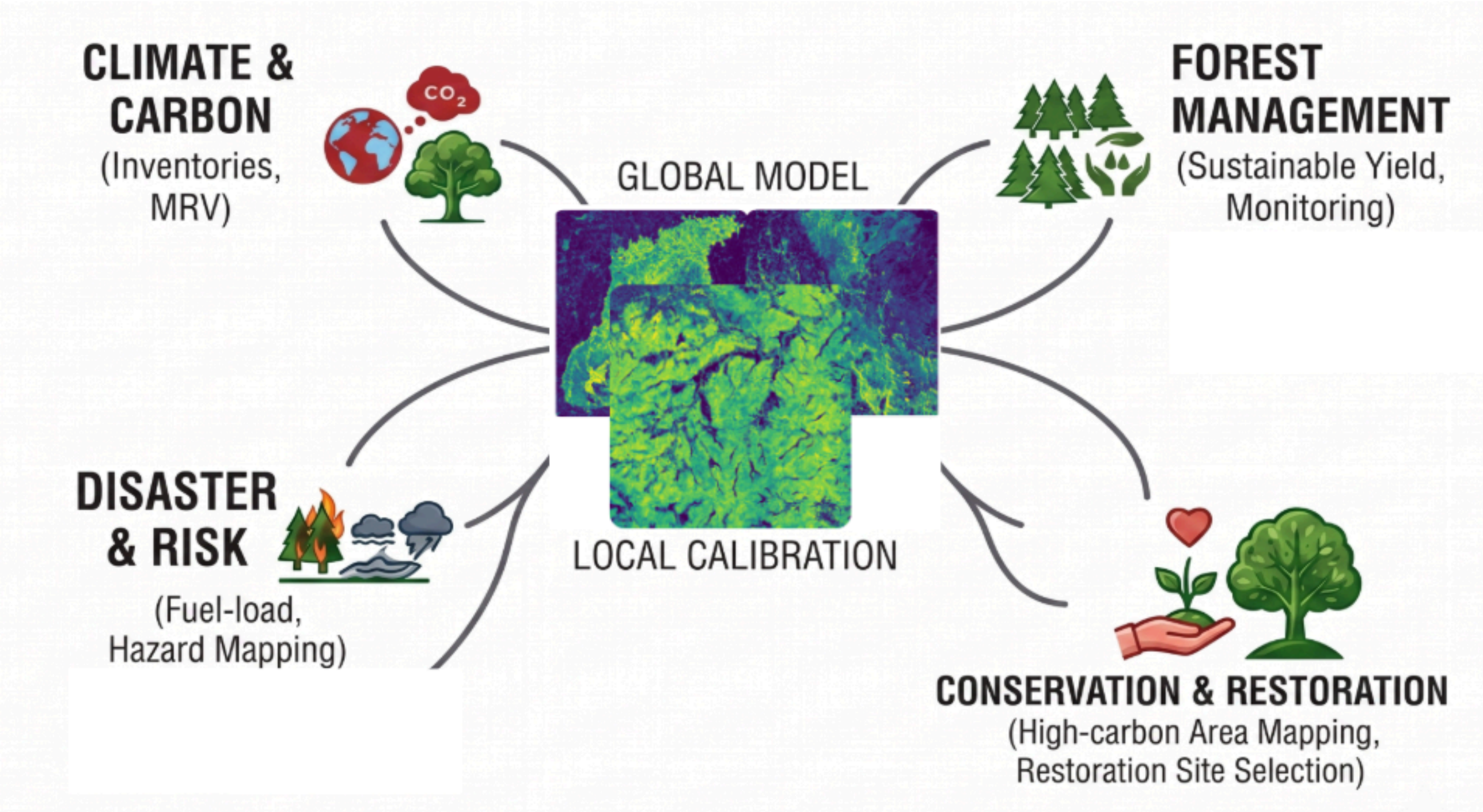

Figure 10. Application domains served by biomass maps.

# 6. DISCUSSION AND CONCLUSION

The results showed that a globally trained model captures the structure biomass relationship well enough to transfer between landscapes. By training a robust deep learning model on a vast, multi-region dataset, the network captures global relationships between multi-sensor structural inputs and aboveground biomass. The subsequent local calibration pipeline combining Random Forest ensemble fine-tuning with a third-order polynomial correction directly targets and removes regional systematic biases. This hybrid design matches or exceeds the predictive accuracy of region-specific models. Even where field data cannot be obtained, the global model remains usable on its own: trained against GEDI reference biomass, it captures the structural signal well enough to produce a meaningful estimate.

However, GEDI footprints do not sample all biomes or slopes equally, and the saturation limits of optical and SAR sensors in high-biomass, dense-canopy forests remain a persistent challenge. Furthermore, while the localized Random Forest fine-tuning and polynomial regression pipeline successfully models non-linear regional trends, the ultimate accuracy of the output depends on the spatial distribution of the local field plots.

In conclusion, this paper delivers a highly portable, operational framework for wall-to-wall aboveground biomass estimation. By combining a globally trained multi-sensor CNN with a lightweight field-calibration workflow, the system establishes a strong global baseline. It also demonstrates exceptional local adaptability: when tested against independent regional inventories, the localized calibration phase elevates the validation performance from an uncalibrated **R2 of 0.11 up to 0.82**, while having the **error** to just **15 Mg/ha**. The outcome is an operational basis for monitoring regional biomass, affordable in its data requirements, and directly serviceable to forest carbon accounting and climate mitigation.

# Acknowledgements

The authors gratefully acknowledge the support of Japan's Ministry of Economy, Trade and Industry (**METI**) through the Subsidies for Global South Future-Oriented Co-Creation Project under the FY2023 Supplementary Budget. We sincerely thank the **Forest Research Institute of Malawi (FRIM)** for providing valuable national field inventory data essential to training and validating the models presented in this study. We are also deeply grateful to **iRise Carbon** for their dedicated efforts in collecting ground-based field inventory data aligned with the national inventory data, which enabled us to confirm our AI model's performance in tracking changes over time.